\documentclass{article}
\usepackage{graphicx}
\usepackage{amsmath,amssymb} 
\usepackage{dsfont}

\usepackage{color}
\usepackage{textcomp}
\usepackage{subfigure}

\newcommand*\samethanks[1][\value{footnote}]{\footnotemark[#1]}

\usepackage[width=122mm,left=12mm,paperwidth=146mm,height=193mm,top=12mm,paperheight=217mm]{geometry}

\title{Human Pose Estimation using Deep Consensus Voting}
\author{Ita Lifshitz\thanks{Equal contribution}, Ethan Fetaya\samethanks\, and Shimon Ullman \\
	Weizmann Institute of Science}
\date{}

\begin{document}

	\maketitle

\begin{abstract}
In this paper we consider the problem of human pose estimation from a single still image. We propose a novel approach where each location in the image votes for the position of each keypoint using a  convolutional neural net. The voting scheme allows us to utilize information from the whole image, rather than rely on a sparse set of keypoint locations. Using dense, multi-target votes,  not only produces good keypoint predictions, but also enables us to compute image-dependent joint keypoint probabilities by looking at consensus voting. This differs from most previous methods where joint probabilities are learned from relative keypoint locations and are independent of the image. We finally combine  the keypoints votes and joint probabilities in order to identify the optimal pose configuration. We show our competitive performance on the MPII Human Pose and Leeds Sports Pose datasets. \\

\end{abstract}

\section{Introduction} 
In recent years,  with the resurgence of deep learning techniques, the accuracy of human pose estimation from a single image has improved dramatically. Yet despite this recent progress, it  is still a challenging computer vision task and state-of-the-art results are far from human performance.\\



The general approach in previous works, such as \cite{pishchulin15,Tompson14}, is to train a deep neural net as a keypoint detector for all keypoints. Given an image $\mathcal{I}$, the net is fed a patch of the image $\mathcal{I}_y\subset \mathcal{I}$  centered around pixel $y$ and predicts if $y$ is one of the $M$ keypoints of the model. This process is repeated in a sliding window approach, using a fully convolutional implementation, to produce $M$ heat maps, one for each keypoint. Structure prediction, usually by a graphical model, is then used to combine these heat maps into a single pose prediction. This approach has several drawbacks. First, most pixels belonging to the person are not themselves any of the keypoints and therefore contribute only limited information to the pose estimation process. Information from the entire person can be used to get more reliable predictions, particularly in the face of partial occlusion where the keypoint itself is not visible. Another drawback is that while the individual keypoint predictors use state-of-the-art classification methods to produce high quality results, the binary terms in the graphical model, enforcing global pose consistency, are based only on relative keypoint location statistics gathered from the training data and are independent of the input image. \\
\begin{figure}[ht]
	\centering
	\subfigure{%
		\includegraphics[width=0.25\textwidth]{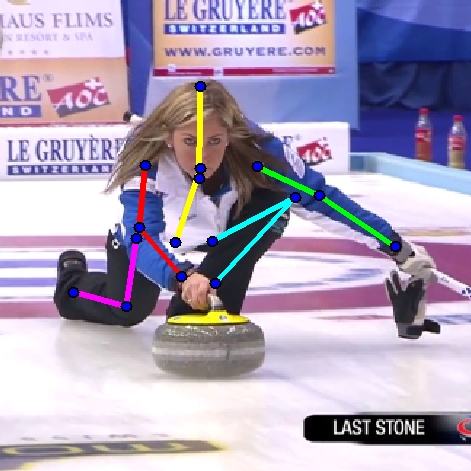}
	}%
	\subfigure{%
		\includegraphics[width=0.25\textwidth]{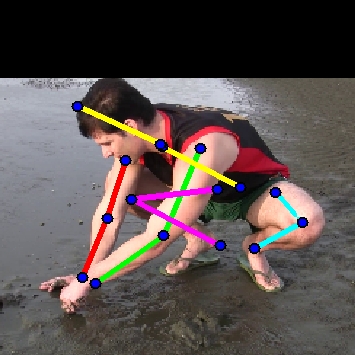}
	}%
	\subfigure{%
		\includegraphics[width=0.25\textwidth]{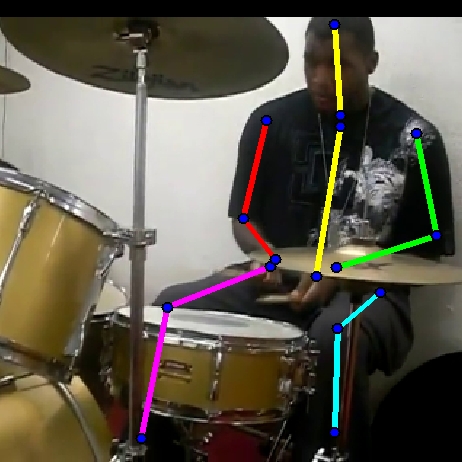}
	}%
	\subfigure{%
		\includegraphics[width=0.25\textwidth]{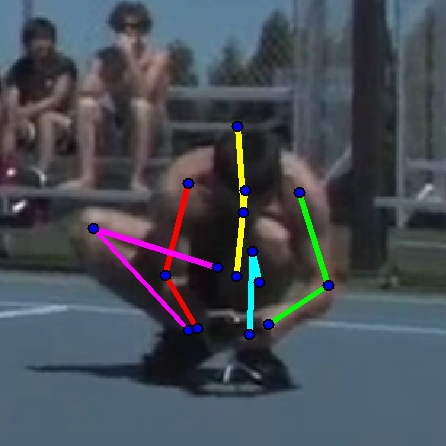}
	}%
	
	\subfigure{%
		\includegraphics[width=0.25\textwidth]{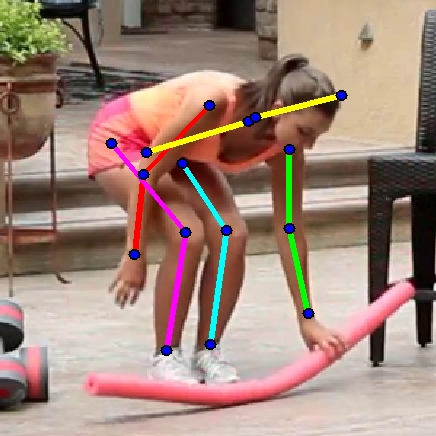}
	}%
	\subfigure{%
		\includegraphics[width=0.25\textwidth]{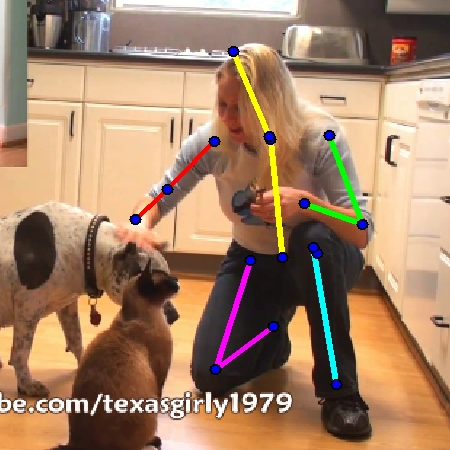}
	}%
	\subfigure{%
		\includegraphics[width=0.25\textwidth]{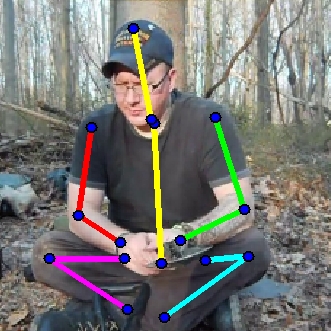}
	}%
	\subfigure{%
		\includegraphics[width=0.25\textwidth]{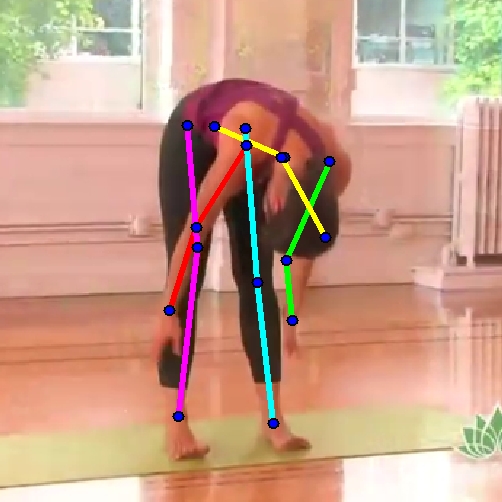}
	}%
	
	\subfigure{%
		\includegraphics[width=0.25\textwidth]{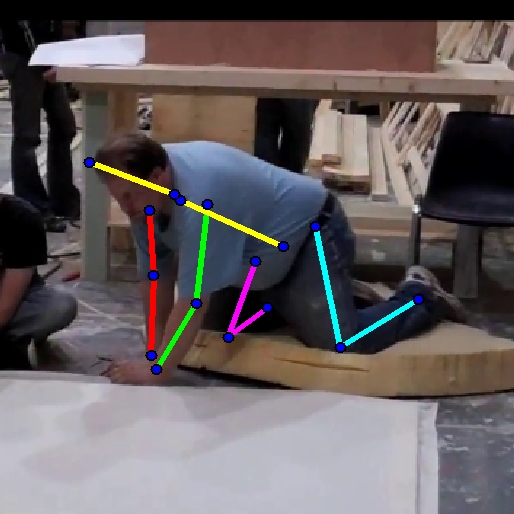}
	}%
	\subfigure{%
		\includegraphics[width=0.25\textwidth]{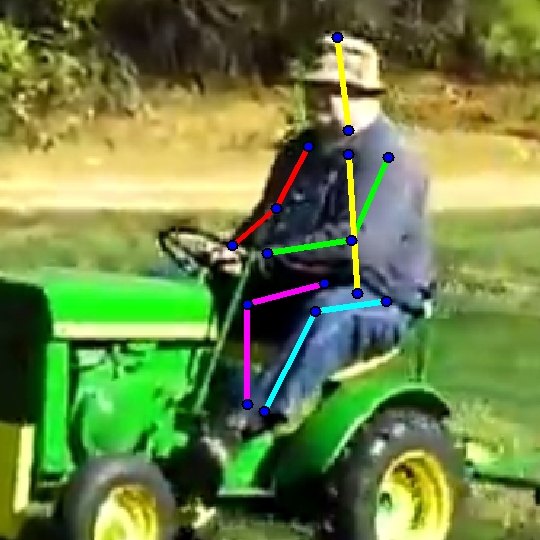}
	}%
	\subfigure{%
		\includegraphics[width=0.25\textwidth]{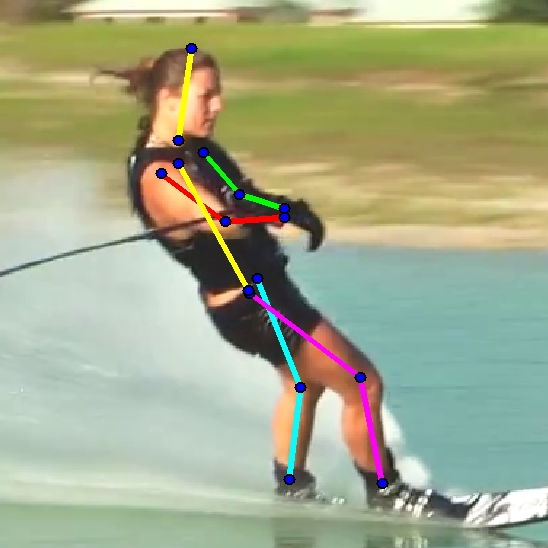}
	}%
	\subfigure{%
		\includegraphics[width=0.25\textwidth]{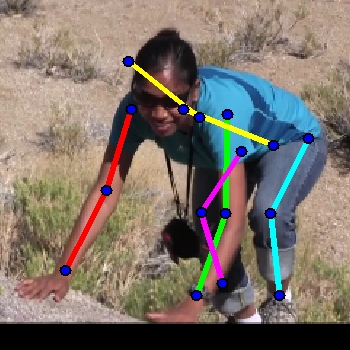}
	}%
	
	\caption{Our model's predicted pose estimation on the MPII-human-pose database test-set \cite{andriluka14}. Each pose is represented as a stick figure, inferred from predicted joints.	Different limbs in the same image are colored differently, same limb across different images has the same color.}
\end{figure}\\
To overcome these limitations, we propose a novel approach in which for every patch center $y$ we predict the location of all keypoints relative to $y$, instead of classifying $y$ as one of the keypoints. This enables us to use 'wisdom of the crowd' by aggregating many votes to produce accurate keypoint detections. In addition, by looking at agreements between votes, we infer informative image-dependent binary terms between keypoints. Our binary terms are generated by consensus voting - we look at a set of keypoints pairs, and for each possible  pair of locations, we aggregate all votes for this combination. The total vote will be high if both keypoint locations get strong votes from the same voters. We show that this approach produces competitive results on the challenging MPII human-pose  \cite{pishchulin13} and the Leeds sports pose \cite{Johnson10} datasets.

\section{Related Work}
Human body pose estimation in still images is a challenging task. The need to cope with a variety of poses in addition to a large range of appearances due to different clothing, scales and light conditions gave rise to many approaches to dealing with the various aspects of this task. The most common approach is to use a keypoint detector combined with a pictorial structure  for capturing relations between parts \cite{felzenszwalb05,yang11}. In addition to the standard pictorial structure \cite{fischler73}, poselet-based features were used to incorporate higher-order part dependencies \cite{pishchulin13}. Alternative methods, like the chains-model  \cite{karlinsky10} replace the pictorial structure with a set of voting chains, starting at the head keypoint and ending at the hand.\\

With the reappearance of convolutional neural nets, part detectors became more reliable, leading to a significant improvement in accuracy  \cite{chen14,Tompson14,carreira15,Tompson15,pishchulin15,wei16}. The works of \cite{Tompson14,Tompson15} focus on multi-scale representation of body parts in order to infer the location of keypoints. In \cite{pishchulin15}, the authors deal with simultaneous detection and pose estimation of multiple people. Recent works \cite{carreira15,wei16} use an iterative scheme to refine pose estimation. As in our approach, in \cite{chen14} an image dependent binary term is  learned. They, however, learn the binary terms explicitly while in our model it arises naturally from the voting scheme.



\section{Overview of the Method}
We now describe the main parts of our algorithm, which will be explained in detail in next sections.\\

At inference, we first use a deep neural net, described in section \ref{sec:net}, to predict for each image patch $\mathcal{I}_y$ centered around pixel $y$, and for each keypoint $j$, the location of the keypoint relative to $y$. From this we can compute the probability of keypoint location $K_j$  being equal to a possible location $x$ as seen from $\mathcal{I}_y$, $P_y(K_j=x)$. We aggregate these votes over all image patches to get the probability distribution for each keypoint location $\{P(K_j=x)\}_{j=1}^M$.  Examples of $P_y(K_j)$ and $P(K_j)$ are shown in figures \ref{fig:vote-wrist}-\ref{fig:vote-shoulder-full} sec. \ref{sec:voting}.\\

Next we compute our consensus voting binary term. The voting net above was trained using a separate loss  per keypoint, which is equivalent to an independence assumption, i.e. for each $y$,
\begin{equation}
P_y(K_i=x_i,K_j=x_j)=P_y(K_i=x_i)\cdot P_y(K_j=x_j).
\end{equation}

If we now average over all locations $y$ we get a joint distribution 
\begin{equation}
P(K_i=x_i,K_j=x_j)\propto \sum_{y} P_y(K_i=x_i)\cdot P_y(K_j=x_j)
\end{equation}
in which the keypoints are no longer independent. Because of the multiplication, the joint distribution is high when both locations get strong votes from the \emph{same voters}. More details on the consensus voting can be found in sec. \ref{sec:consensus}.\\

Finally, we estimate the pose by minimizing an energy function over the unary and binary terms generated by the voting scheme. We do this sequentially, focusing at each step on a subset of keypoints. We start with the most reliable keypoints until making the full pose estimation. This process is presented in more details in sec. \ref{sec:pose-prediction}.

\section{Keypoint Voting}\label{sec:voting}
\subsection{Voting Representation}

The first stage in our pose-estimation method is a keypoint detector learned by a deep neural net, which we apply to the image patches in a fully convolutional fashion \cite{Long15}. This is an accelerated way to run the net in a sliding window manner with the stride determined by pooling layers of the network. At each patch center $y$, the net predicts the location of each keypoint $K_1,...,K_M$ relative to $y$. This differs from previous methods \cite{Tompson14,Tompson15} where the net only needed to classify if the center $y$ is any of the $M$ keypoints.\\

\begin{figure}[ht]
	\centering
	\subfigure[Log-polar bins]{%
		\includegraphics[width=0.3\textwidth]{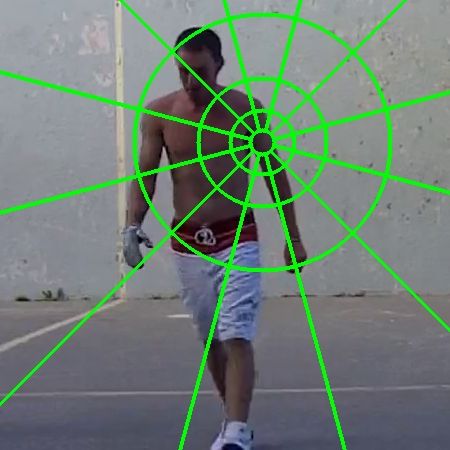}
		\label{fig:log-rad-grid}}
	\subfigure[Ignore mask]{%
		\includegraphics[width=0.3\textwidth]{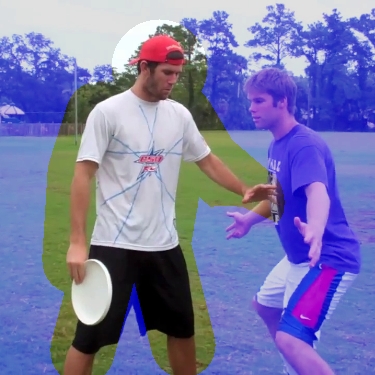}
		\label{fig:ignore-mask}}
	\subfigure[Keypoints]{%
		\includegraphics[width=0.235\textwidth]{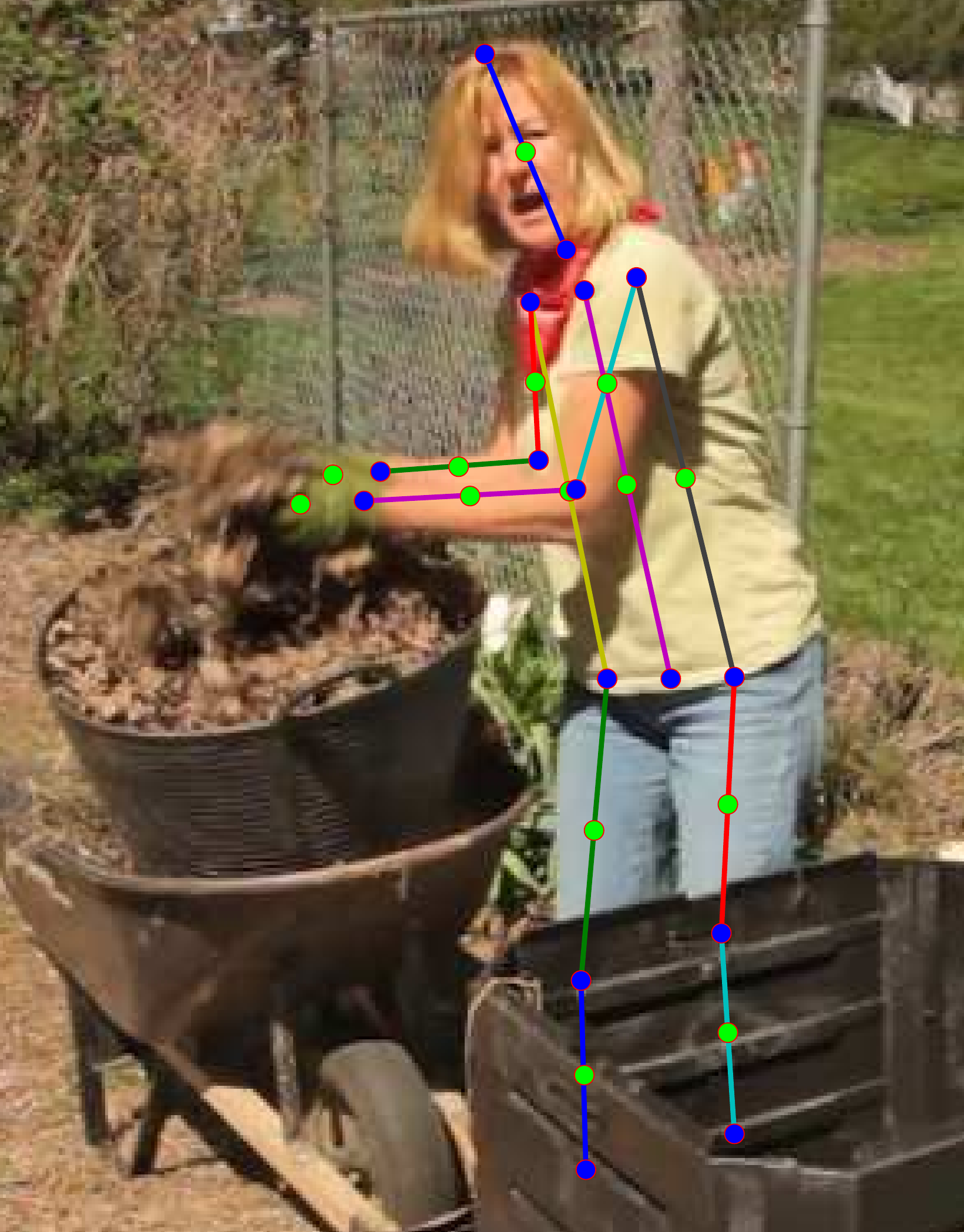}
		\label{fig:gt}}
	\caption{(a) Log-polar bins, used for keypoints locations voting, centered around the left upper arm; (b) Patch centers outside the person of interest, marked in blue, are not used for training; (c) Our model makes use of 30 keypoints: 16 annotated body joints, supplied in the dataset, 12 synthetically generated mid-section keypoints and estimated hands locations. Original keypoints marked in blue, synthetically generated keypoints in green }
	\label{fig:intro}
\end{figure}

The problem of predicting the relative displacement vectors $\{K_1-y,...,K_M-y\}$ is a regression problem which is usually solved by minimizing an $L_2$ loss function. However, for the current task the $L_2$ loss has shortcomings, as the net produces only a single prediction as output. Consequently, in cases of ambiguity, e.g. difficulty to distinguish between left and right hand, the optimal $L_2$ loss would be the mean location instead of ``splitting'' the vote between both locations. Indeed, when trained using this approach,  we found that the net performance is degraded by this problem. To better address this issue, we modify the prediction task into a classification problem by discretizing the image into log-polar bins centered around $y$, as seen in Fig. \ref{fig:log-rad-grid}. Using log-polar binning allows for more precise predictions for keypoints near $y$ and a rough estimate for far away keypoints. We classify into 50 classes, one for the central location, one for background i.e. non-person, and the rest are four log-polar rings around the center with each ring divided into 12 angular bins. Since not all people in the training set are labeled, we are unable to use background locations for training the non-person label. For this reason, we ignore image locations far from the person of interest, as seen in Fig. \ref{fig:ignore-mask}, and use non-person images from the PASCAL dataset for non-person label training. \\ 

We augmented the 16 humanly annotated keypoints supplied in the dataset with additional 12 keypoints generated from the original ones, by taking the middle point between two neighboring skeleton keypoints, e.g. the middle point between the shoulder and elbow. We also obtained estimated location of the hands by extrapolating the elbow-wrist vector by 30\%. This produces a total of 30 keypoints and allows us to have a more dense coverage of the body. All keypoints can be seen in Fig. \ref{fig:gt}. \\


\subsection{Net Architecture}\label{sec:net}

\begin{table}[ht]
	\centering
	\resizebox{\textwidth}{!}{\begin{tabular}{r *{14}{|c}} \hline
			\multicolumn{15}{c}{(a) $VGG_{16}$: 244x244x3 input image; 1x1000 output labels } \\
			\hline
			& 1 & 2 & 3 & 4 & 5 & 6 & 7 & 8 & 9 & 10 & 11 & 12 & 13 & \\ \hline
			\textit{layer} & 2 x conv & max & 2 x conv & max & 3 x conv & max & 3 x conv & max & 3 x conv & max & fc & fc & fc & \\
			\textit{filter-stride} & 3-1 & 2-2 & 3-1 & 2-2 & 3-1 & 2-2 & 3-1 & 2-2 & 3-1 & 2-2 & - & - & - & \\
			\textit{channels} & 64 & 64 & 128 & 128 & 256 & 256 & 512 & 512 & 512 & 512 & 4096 & 4096 & 1000  & \\
			\textit{activation} & relu & idn & relu & idn & relu & idn & relu & idn & relu & idn & relu& relu & soft & \\
			\textit{size} & 224 & 112 & 112 & 56 & 56 & 28 & 28 & 14 & 14 & 7 & 1 & 1 & 1 & \\ \hline
			\multicolumn{15}{c}{(b) $HPE$-$WIS$: 504x504x3 input image; 102x102x50x30 output label maps } \\
			\hline
			& 1 & 2 & 3 & 4 & 5 & 6 & 7 & 8 & 9 & 10 & 11 & 12 & 13 & 14 \\ \hline
			\textit{layer} & $2\times conv$ & $max$ & $2\times conv$ & $max$ & $3\times conv$ & $max$ & $3\times conv$ & $max$ & $3\times hconv$ & $max $ & $hconv$ & $conv$ & $conv_{K_i}$ & $deconv_{K_i}$\\
			\textit{filter-stride} & 3-1 & 2-2 & 3-1 & 2-2 & 3-1 & 2-2 & 3-1 & 2-1 & 3-1 & 3-1 & 7-1 & 1-1 & 1-1 & 6-2 \\
			\textit{channels} & 64 & 64 & 128 & 128 & 256 & 256 & 512 & 512 & 512 & 512 & 2048 & 2048 & 50 & 50 \\
			\textit{activation} & relu & idn & relu & idn & relu & idn & relu & idn & relu & idn & relu & relu & idn & w-soft \\
			\textit{size} & 504 & 252 & 252 & 126 & 126 & 63 & 63 & 62 & 62 & 60 & 50 & 50 & 50 & 102 \\ \hline 
			\multicolumn{15}{c}{ } \\
		\end{tabular}}
		\caption{Comparisons between the network architectures of $VGG_{16}$ and $HPE$-$WIS$, as shown in (a) and (b). Each table contains five rows, representing the \textquotesingle name of layer\textquotesingle, \textquotesingle receptive field of filter \-- stride\textquotesingle, \textquotesingle number of output feature maps\textquotesingle , \textquotesingle activation function\textquotesingle \, and \textquotesingle size of output feature maps\textquotesingle , respectively. The terms \textquotesingle conv\textquotesingle, \textquotesingle max\textquotesingle, \textquotesingle fc\textquotesingle, \textquotesingle hconv\textquotesingle \, and \textquotesingle deconv\textquotesingle \,  represent the convolution, max pooling, fully connection, convolution with holes \cite{chen2014semantic} and deconvolution upsampling \cite{Long15}, respectively. The terms \textquotesingle relu\textquotesingle, \textquotesingle idn\textquotesingle, \textquotesingle soft\textquotesingle \, and \textquotesingle w-soft\textquotesingle \, represent the activation functions, rectified linear unit, identity, softmax and weighted-softmax, respectively. The last two layers are distinct per keypoint.} 
		\label{table:net}
	\end{table} 
	
	The net architecture we use is based on the VGG-16 network \cite{Simonyan15}. We use the first 13 convolutional layers parameters of the VGG-16 net, pre-trained on the imagenet dataset \cite{russakovsky15}. On top of these layers we employ a max pooling with stride 1 and use convolution with holes \cite{chen2014semantic} in order to increase resolution and generate a descriptor of size 2048 every 8 pixels. To further increase resolution we learn an upsampling deconvolution layer  \cite{Long15} and get a probability distribution over 50 classes, indicating the relative keypoint location, every 4 pixels. The last two layers are distinct per keypoint resulting in $30$ distributions on $50$ bins every $4$ pixels. More details about the structure of the net can be found in Table \ref{table:net}. The net training was done over images from the MPII Human pose dataset \cite{andriluka14} in a cascaded fashion. First, training the added layers, denoted by layers 11-14, in table \ref{table:net}(b), while keeping the first 13  convolutional layers with learning rate 0. Then training layer 9 in table \ref{table:net}(b) and finally fine tuning the entire network. We start with learning rate of 0.001 and no weight decay and continue with learning rate of 0.0001 and 0.0002 weight decay. Since the size of  each classification bin is considerably different, we minimize a weighted softmax loss function, where each class is weighted inversely proportional to the size of its bin.   \\
	
	\subsection{The Voting Scheme}\label{sec:scheme-voting}
\begin{figure}[ht]
		\centering
		\begin{minipage}{.65\textwidth}
			\subfigure[Right wrist single vote map]{%
				\includegraphics[width=0.45\textwidth]{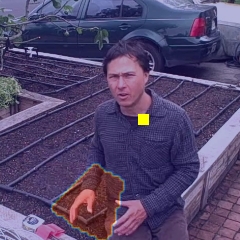}
				\label{fig:vote-wrist}}
			\quad
			\subfigure[Left shoulder single vote map]{%
				\includegraphics[width=0.45\textwidth]{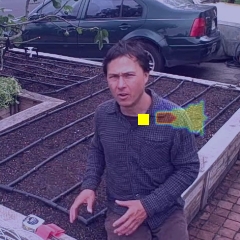}
				\label{fig:vote-shoulder}}
			
			\subfigure[Right wrist probability]{%
				\includegraphics[width=0.45\textwidth]{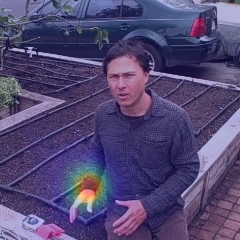}
				\label{fig:vote-wrist-full}}
			\quad
			\subfigure[Left shoulder probability]{%
				\includegraphics[width=0.45\textwidth]{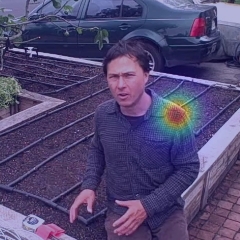}
				\label{fig:vote-shoulder-full}}
		\end{minipage}
		\begin{minipage}{.3\textwidth}
			\subfigure[deconvolution kernel]{%
				\includegraphics[width=1\textwidth]{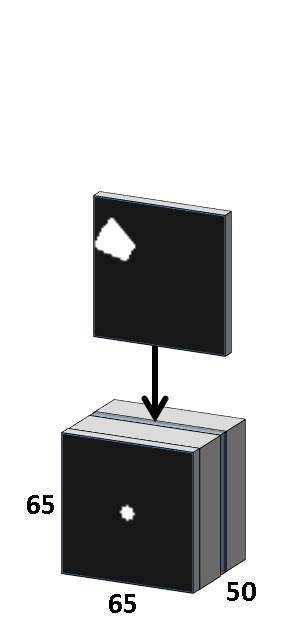}
				\label{fig:deconv-kernel}}
		\end{minipage}
		\caption{(a) Voting map from location y (yellow rectangle) for Right Wrist $P_y(K_{Right Wrist})$; (b) Voting map from location y (yellow rectangle) for Left Shoulder  $P_y(K_{Left Shoulder})$; (c) Right wrist probability $P(K_{Right Wrist})$, generated by aggregating voting maps for right wrist; (d) Left shoulder probability $P(K_{Left Shoulder})$, generated by aggregating voting maps for left shoulder; (e) The deconvolution kernel $w$. The weights of a specific channel are illustrated at the top}
		\label{fig:single-vote}
	\end{figure}

	At each patch center $y$ and for each keypoint $j$ the network returns a softmax probability distribution over the log-polar bins  $s^{j}_y\in \mathds{R}^{1\times 1\times C}$. At inference we use deconvolution, with a predefined fixed kernel $w\in \mathds{R}^{H_k\times W_k\times C}$, to transform the log-polar representation $s^{j}_y$ back to the image space and get the probability distribution of keypoint location over pixels $P_y(K_j=x)$. The deconvolution kernel maps each channel, representing a log-polar bin, to the corresponding image locations. We use a deconvolution kernel of size ${(H_k\times W_k\times C)}$ $=$ ${(65\times 65\times 50)}$. Most of the kernel weights are zeros, shown as black in fig. \ref{fig:deconv-kernel}. At the top of the figure we show an illustration of weights for a specific bin. Since this bin corresponds to the upper left log-polar segment it is zero at all locations except for the pixels of that segment which are set to $\frac{1}{|bin|}$.

	\begin{equation}
	\hat{P_y}(K_j=x)=deconv(s^{j}_y,w)
	\end{equation}
	
	Then $P_y(K_j)$ is simply $\hat{P_y}(K_j)$ translated by y. Examples for $P_y(K_j)$ are shown in fig. \ref{fig:vote-wrist} and \ref{fig:vote-shoulder}. We aggregate these votes over all patch centers to get the final probability of each keypoint at each location.
	
	\begin{equation}
	P(K_j=x)=\sum_{y\in \mathcal{Y}}P_y(K_j=x)=deconv(s^{j},w)
	\end{equation}
	
	The term $P(K_j)\in \mathds{R}^{(\frac{H}{4}+H_k-1)\times (\frac{W}{4}+W_k-1)}$ is the aggregated votes of keypoint $j$, and $s^{j}\in \mathds{R}^{\frac{H}{4}\times \frac{W}{4}\times C}$ is the softmax distribution output of keypoint $j$. Examples for $P(K_j)$ are shown in fig. \ref{fig:vote-wrist-full} and \ref{fig:vote-shoulder-full}. Note that the size of the aggregated distribution is larger than $\frac{H}{4}\times \frac{W}{4}$, the image size at the output resolution. This enables the algorithm to generate votes for locations outside the visible image.\\
	
	
	During inference, we make use of the Caffe \cite{jia2014} implementation for deconvolution by adding an additional deconvolution layer with fixed weights $w$ on top of the net softmax output. This layer generates the aggregated probability distribution of each keypoint in a fast and efficient way.



\section{Consensus Voting}\label{sec:consensus}
The voting scheme described in section \ref{sec:voting} produces a heat map per keypoint, which represents the probability of a keypoint being at each location. The next challenge is to combine all the estimated distributions into a single pose prediction. What makes this task especially challenging is that beyond just having to handle false keypoint detections, we need to handle true detections but of other individuals in the image as well. Other challenges include self-occlusion, and confusion between left and right body parts.   \\

The standard approach to solve this problem is to minimize an objective of the following form:
\begin{equation}\label{eq:binary}
\sum_{i=1}^N\phi_i(x_i)+ \sum_{(i,j)\in E}\phi_{(i,j)}(x_i,x_j)
\end{equation}
The $\phi_i(x_i)$ unary term is the score of having keypoint $i$ at location $x_i$. The  $\phi_{(i,j)}(x_i,x_j)$ binary term measures compatibility and is the score of having keypoint $i$ at location $x_i$ and keypoint $j$ at location $x_j$. The edge set $E$ is some preselected subset of pairs of keypoints.\\

Usually in pose estimation, the binary term is image independent and comes from a prior on relative keypoint location. For example, given the location of the shoulder, it produces a set of possible elbow displacements based on training samples pose statistics. While this can be helpful in excluding anatomically implausible configurations, an image independent binary term has limitations. For example, the location of the left shoulder relative to the head is strongly dependent on whether the person is facing forwards or backwards and this information is not incorporated into the standard binary term. One main advantage of our keypoint voting scheme is that we can compute from it an image-based ``consensus voting'' binary term, which has more expressive power. This is especially important for less common poses, where the image-independent prior gives a low probability to the right pose. \\

\begin{figure}[ht]
	\centering
	\subfigure[Left shoulder probability]{%
		\includegraphics[width=0.35\textwidth]{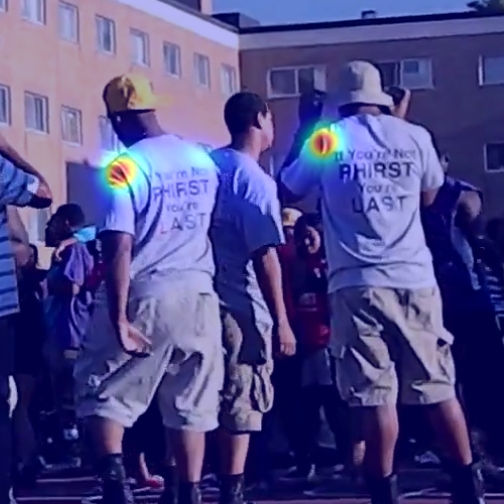}
		\label{fig:conditional-probability-1}}
	\quad
	\subfigure[Left elbow probability]{%
		\includegraphics[width=0.35\textwidth]{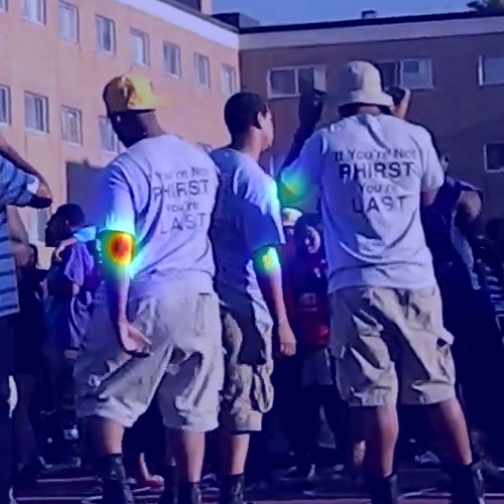}
		\label{fig:conditional-probability-2}}
	
	\subfigure[Conditional probability]{%
		\includegraphics[width=0.35\textwidth]{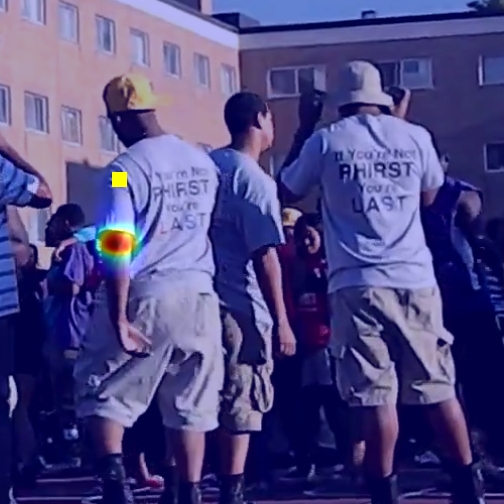}
		\label{fig:conditional-probability-3}}
	\quad
	\subfigure[Conditional probability]{%
		\includegraphics[width=0.35\textwidth]{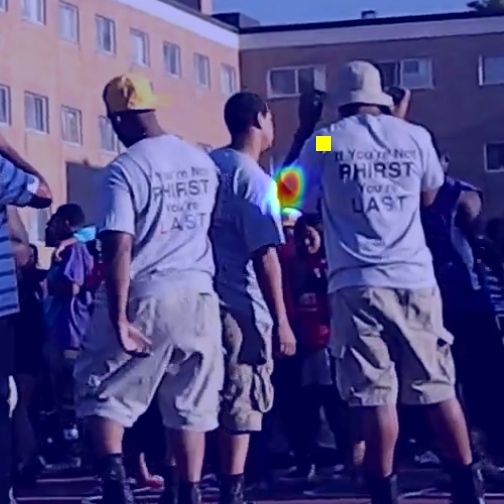}
		\label{fig:conditional-probability-4}}
	\caption{(a) Left shoulder probability $P(K_{Left Shoulder})$; (b)Left elbow probability $P(K_{Left Elbow})$; (c) Conditional probability of left elbow given left shoulder of person on the left (yellow rectangle); (d) Conditional probability of left elbow given left shoulder of person on the right (yellow rectangle)}
	\label{fig:conditional-probability-all}
\end{figure}

At each image location $y$, we compute from the net output the distribution $P_y(K_i=x)$ for each keypoint. These probabilities were trained under a (naive) independence assumption, i.e. a separate softmax loss for each keypoint, more formally 
\begin{equation}
P_y(K_i=x_i,K_j=x_j)=P_y(K_i=x_i)\cdot P_y(K_j=x_j).
\end{equation}

If we now average over all locations $y$, we get a joint distribution 
\begin{equation}
P(K_i=x_i,K_j=x_j)\propto \sum_{y} P_y(K_i=x_i)\cdot P_y(K_j=x_j)
\end{equation}\label{eq:objective}

which is no longer independent. By having each center $y$ vote for a combination of keypoints, the probabilities become dependent through the voters. For $P(k_i=x_i,k_j=x_j)$ to be high, it is not enough for $x_i$ and $x_j$ to receive strong votes separately, the combination needs to get strong votes from many common voters. For this reason we call this the consensus voting term.\\

In Fig. \ref{fig:conditional-probability-all} we show an example of the conditional probability $P(K_i=x_i|K_j=x_j)$ calculated from the previously described joint probability. As can be seen in Fig. \ref{fig:conditional-probability-2}, the left elbow of the  person to the right has a weak response. In addition, there is a misleading weak response to the elbow of another individual nearby. After conditioning on the left shoulder location, we see  a strong response in Fig \ref{fig:conditional-probability-4}, but only in the correct location.\\

In our algorithm, we use the unary term of the form $\phi_i(x_i)=-\log(P(K_i=x_i))$. The binary term $\phi_{(i,j)}(x_i,x_j)$ we use is a weighted combination of the joint distribution $-\log(P(k_i=x_i,k_j=x_j))$ just described and the commonly used binary term based on relative keypoint location statistics. The hyperparameters where tuned using the TRW-S algorithm \cite{Kolmogorov06} on a validation set.\\

Computing the consensus voting is challenging, since calculating equation 7 in a naive way is computationally expensive. Each keypoint has $N^2$ possible locations, where the image is of size $N\times N$. Considering all possible combinations yields $N^4$ pairs of locations. For each pair of possible locations, we sum over all $N^2$ voters, resulting in $\mathcal{O}(N^6)$ running time. In order to reduce running time, we use the observation that eq. 7 is in fact a convolution and use the highly optimized Caffe GPU implementation \cite{jia2014} to calculate the binary tables. In addition we calculate the binary term over a coarse scale of 1/12 of the image scale, using only the first two log-polar rings. This reduces the running time of a single keypoint pair to $\sim 100ms$.   The restriction to the first two rings limits the maximal distance between keypoints, which we overcome by using the augmented keypoints shown in Fig \ref{fig:gt}.

\section{Pose Prediction}\label{sec:pose-prediction}

In the previous sections, we described the unary and binary terms which are the basic building blocks of our algorithm. We now present additional steps  that we employ  to improve performance on top of these basic building blocks. First, we add geometrical constraints on our augmented keypoints. Second, we perform the inference in parts, starting from the reliable parts proceeding to the less certain ones.


\subsection{Local Geometric Constraints}
We generate additional keypoints, as seen in Fig. \ref{fig:gt} by taking the mid-point between keypoints, e.g. shoulder and elbow. While we could simply minimize eq. \ref{eq:objective} with these added variables as well, this fails to take into account  the fact that these new points are determined by other points. We can enforce these constraints by removing the new synthetic variables and rewriting our binary constraints. Assume our two original keypoints had indexes $i$ and $j$ and the middle point had index $\ell$. Focusing on the relevant terms, instead of solving 

\begin{equation}
\min_{x_i,x_j,x_\ell} \phi_i(x_i)+\phi_j(x_j)+\phi_\ell(x_\ell)+\phi_{(i,\ell)}(x_i,x_\ell)+\phi_{(\ell,j)}(x_\ell,x_j) 
\end{equation}
we  add the geometric constraint by solving

\begin{equation}
\min_{x_i,x_j} \phi_i(x_i)+\phi_j(x_j)+\tilde{\phi}_{(i,j)}(x_i,x_j).
\end{equation}

Where we define 
\begin{equation*}
\tilde{\phi}_{(i,j)}(x_i,x_j) = \phi_\ell(f(x_i,x_j))+\phi_{(i,\ell)}(x_i,f(x_i,x_j))+\phi_{(\ell,j)}(f(x_i,x_j),x_j)
\end{equation*}
and $f(x,y) = \frac{1}{2}(x+y)$. \\

This is equivalent to adding the constraint that $x_\ell$ is the  mid-point between $x_i$ and $x_j$, but faster to optimize. By adding this mid-point constraint, and using it as a linking feature \cite{karlinsky12}, we get a more reliable binary term which also looks at the appearance of the space between the two respective keypoints.

\subsection{Sequential Prediction} \label{sec:seq}
An issue that arises when optimizing eq. \ref{eq:objective} over all keypoint is that not all keypoints are detected with equal accuracy. Some, like the head, are detected with high accurately while others, like the wrist, are more difficult to locate. In some cases, occluded or unclear keypoints can distort predictions of more visible keypoints. In order to have the more certain keypoints influence the prediction of less certain ones, but not vice versa, we predict the keypoints in stages. 
We start with the most reliable keypoints, and at each stage use the previously predicted keypoints as an "anchor" to predict the other parts. We have three stages. First, we locate the head, neck, thorax and pelvis. After that we locate the shoulders and hips. Last, we locate all remaining keypoints.

\section{Results}\label{sec:res}
\subsection{MPII}
We tested our method on the MPII human pose dataset \cite{andriluka14}, which consists of 19,185 training and 7,247 testing images of various activities containing over $40K$ annotated people. The dataset is highly challenging and has people in a wide array of poses. At test time we are given an image with a rough location and scale of the person of interest and need to return the location of 16 keypoints: head-top, upper-neck, thorax, pelvis, shoulders, elbows, wrist, hips, knees and ankles.\\

\begin{figure}[!ht]
	\centering
	\subfigure{%
		\includegraphics[width=0.2\textwidth]{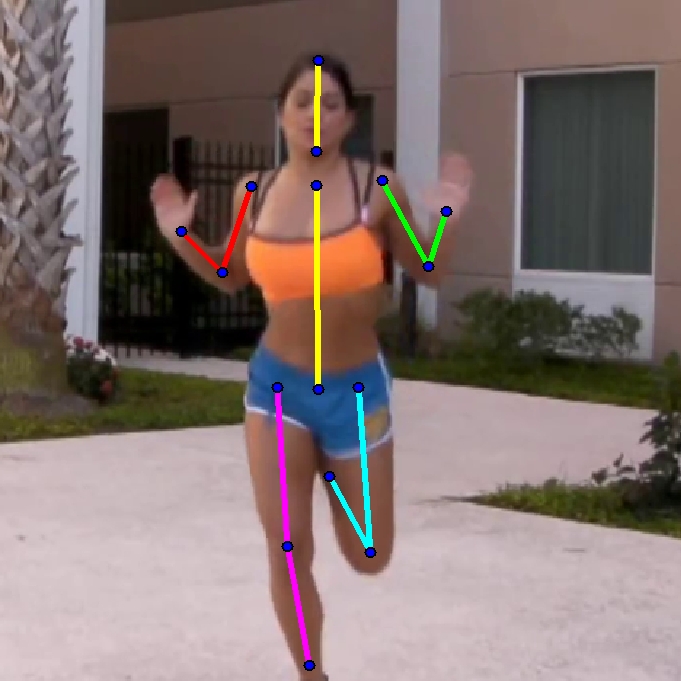}
	}%
	\subfigure{%
		\includegraphics[width=0.2\textwidth]{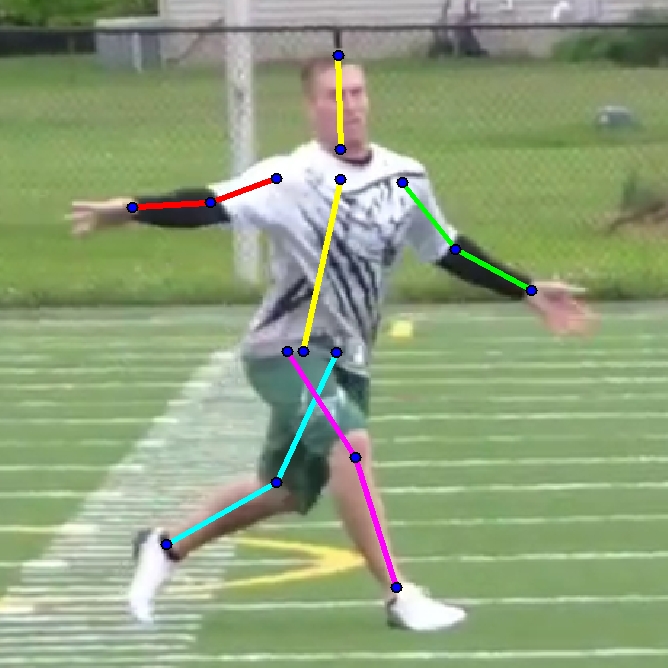}
	}%
	\subfigure{%
		\includegraphics[width=0.2\textwidth]{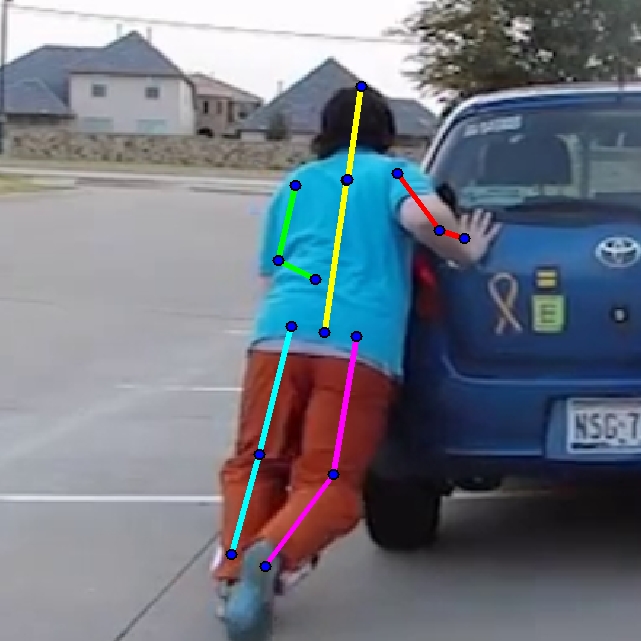}
	}%
	\subfigure{%
		\includegraphics[width=0.2\textwidth]{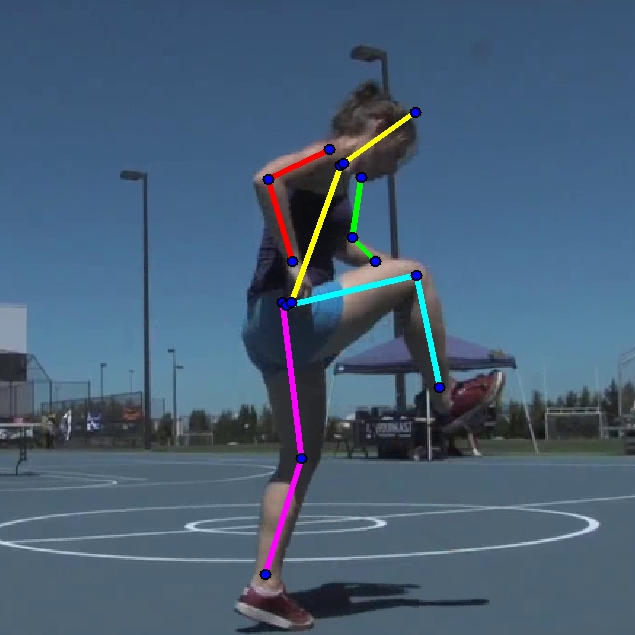}
	}%
	\subfigure{%
		\includegraphics[width=0.2\textwidth]{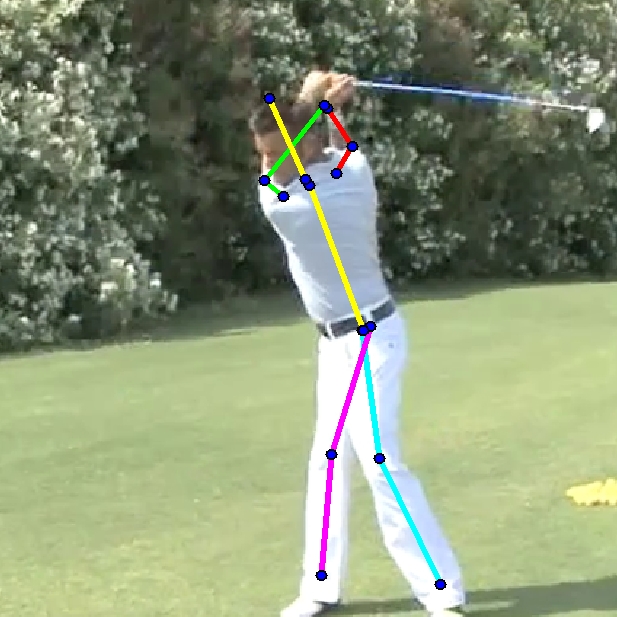}
	}%
	
	\subfigure{%
		\includegraphics[width=0.2\textwidth]{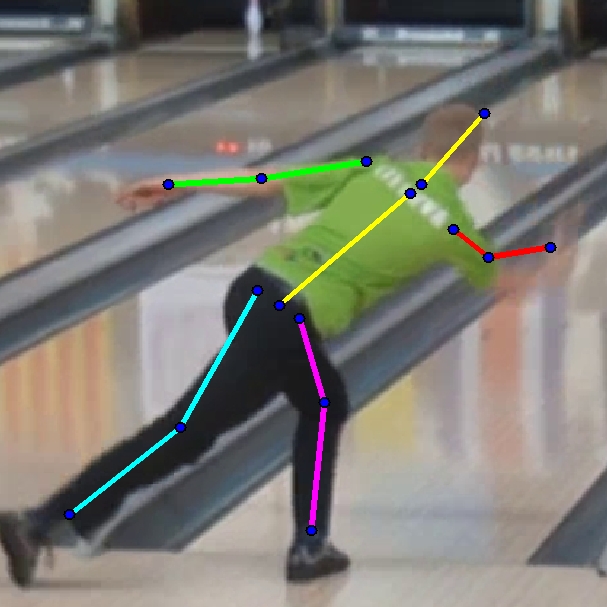}
	}%
	\subfigure{%
		\includegraphics[width=0.2\textwidth]{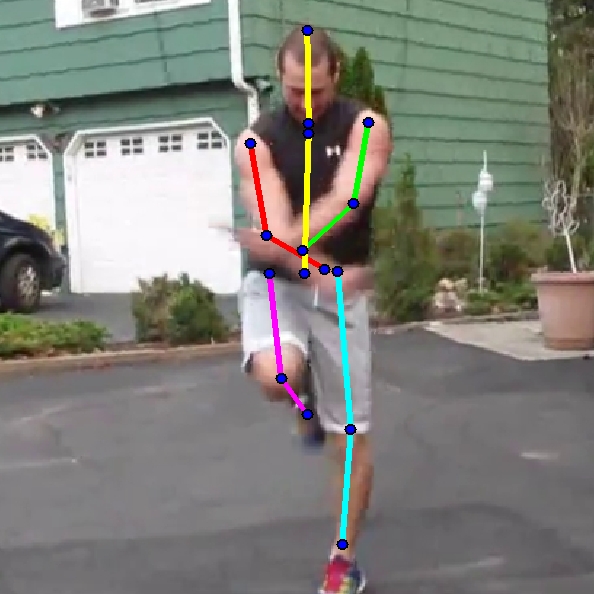}
	}%
	\subfigure{%
		\includegraphics[width=0.2\textwidth]{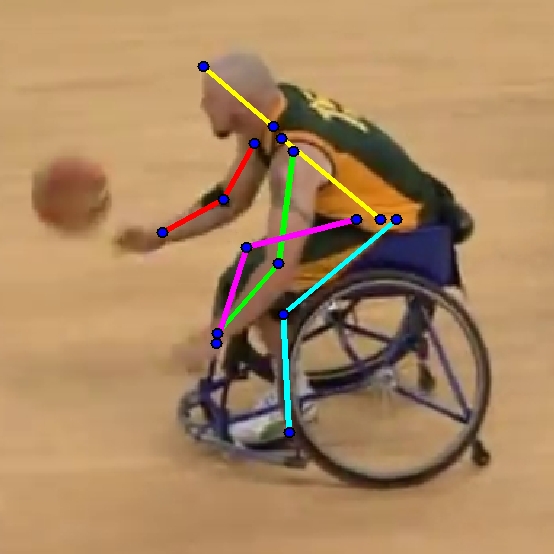}
	}%
	\subfigure{%
		\includegraphics[width=0.2\textwidth]{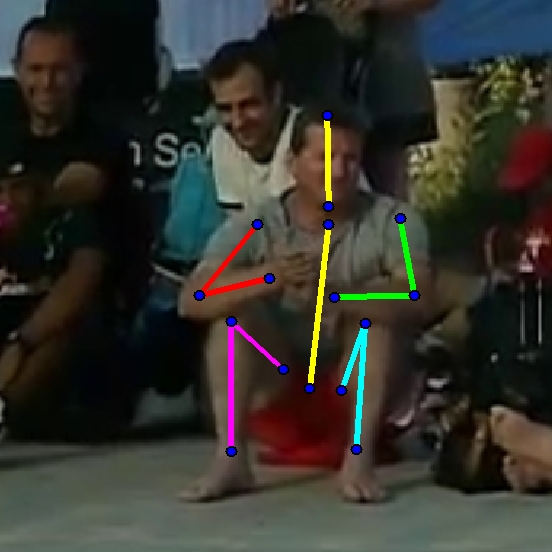}
	}%
	\subfigure{%
		\includegraphics[width=0.2\textwidth]{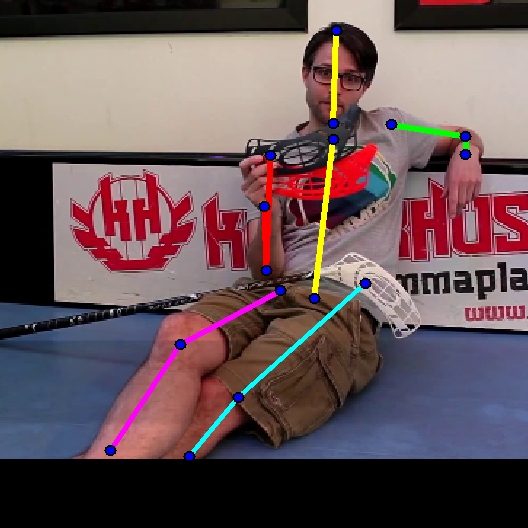}
	}%
	
	\caption{Additional results of our model's predicted joint positions on the MPII-human-pose database test-set \cite{andriluka14}}
	\label{fig:more-mpii-pred-images}
\end{figure}

The standard evaluation is made on a subset of test images, named "single person", where the person of interest is well separated from other people. We note that several images in the "single person" dataset still have another person nearby. In order to restrict our algorithm to the person in question, we crop a window of size $504\times504$ (using zero padding if needed) around the person using the given position and scale. In addition, to insure that we return the pose of the correct person, we  multiply our mid-body keypoint heatmap (the synthetic point between thorax and pelvis) with a mask centered around the given person position. We trained the net described in section \ref{sec:voting} using Caffe \cite{jia2014} and get the final keypoint prediction by sequential optimization, as described in section \ref{sec:seq}, using the TRW-S algorithm \cite{Kolmogorov06}. Various hyperparameters where tuned using a validation set containing 3300 annotated poses. Examples of our model's predictions are shown in fig. \ref{fig:more-mpii-pred-images}.



\begin{table}[!h]
	\centering
	
	\label{my-label}
	\resizebox{\linewidth}{!}{%
	\begin{tabular}{|l|l|l|l|l|l|l|l|l|}
		\hline
		\textbf{Method} & \textbf{Head} & \textbf{Shoulder} & \textbf{Elbow} & \textbf{Wrist} & \textbf{Hip} & \textbf{Knee} & \textbf{Ankle} & \textbf{Mean} \\ \hline
		
		Tompson et al.\cite{Tompson14} & 95.8  & 90.3  & 80.5  & 74.3  & 77.6  & 69.7 & 62.8 & 79.6 \\ \hline
		Carreira et al.\cite{carreira15}\footnotemark[1]& 95.7  & 91.7  & 81.7  & 72.4  & 82.8  & 73.2 & 66.4 & 81.3 \\ \hline
		Tompson et al.\cite{Tompson15}& 96.1  & 91.9  & 83.9  & 77.8  & 80.9  & 72.3 & 64.8 & 82.0\\ \hline
		Pishchulin et al.\cite{pishchulin15}\footnotemark[1]& 94.1  & 90.2  & 83.4  & 77.3  & 82.6  & 75.7 & 68.6 & 82.4  \\ \hline
		Wei et al.\cite{wei16}\footnotemark[1] & 97.7  & \bf{94.5}  & \bf{88.3}  & \bf{83.4}  & \bf{87.9}  & \bf{81.9} & \bf{78.3} & \bf{87.9}
		\\ \hline
		Our Model& \bf{97.8}  & 93.3  &  85.7  & 80.4  & 85.3  & 76.6 & 70.2 & 85.0 \\ \hline
		
	\end{tabular} }
	\caption{PCKh results on the MPII single person dataset. Works marked with $\star$ are arXIV preprints, not yet peer-reviewed.}
	
\end{table}

Performance is measured by the PCKh metric \cite{andriluka14}, where a keypoint location is considered correct if its distance to the ground truth location is no more than half the head segment length. In table 2 we compare our results to the leading methods on the MPII human pose leaderboard. We show competitive results with mean PCKh score of $85.0\%$ and state-of-the-art performance on the head keypoints.

\begin{figure}[ht]
	\centering
	\subfigure{%
		\includegraphics[width=0.33\textwidth,
		height=0.3\textwidth]{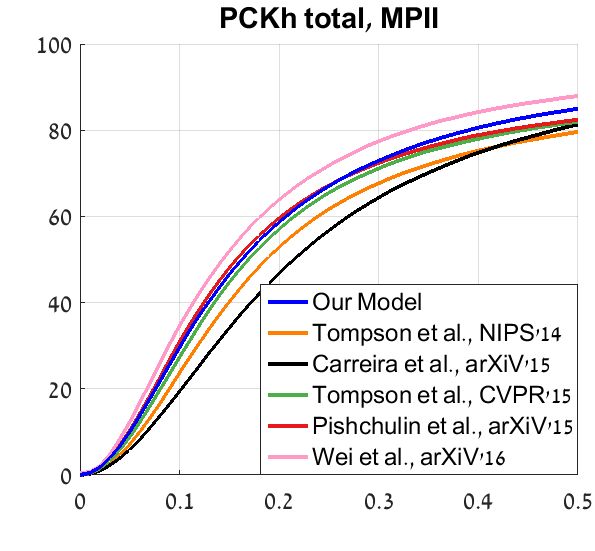}
	}%
	\subfigure{%
		\includegraphics[width=0.33\textwidth,
		height=0.3\textwidth]{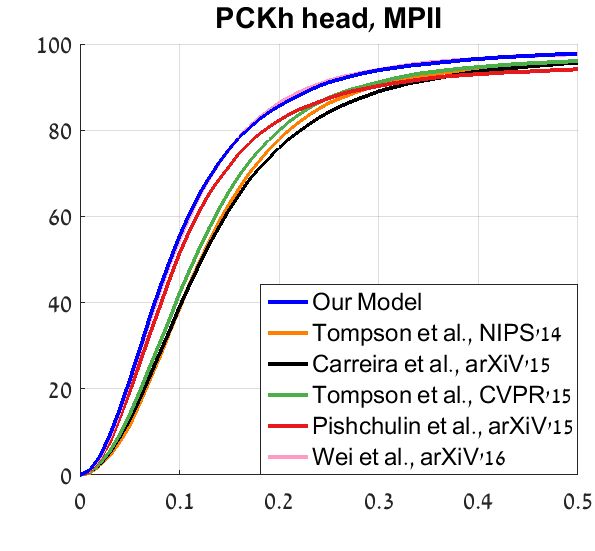}
	}%
	\subfigure{%
		\includegraphics[width=0.33\textwidth,
		height=0.3\textwidth]{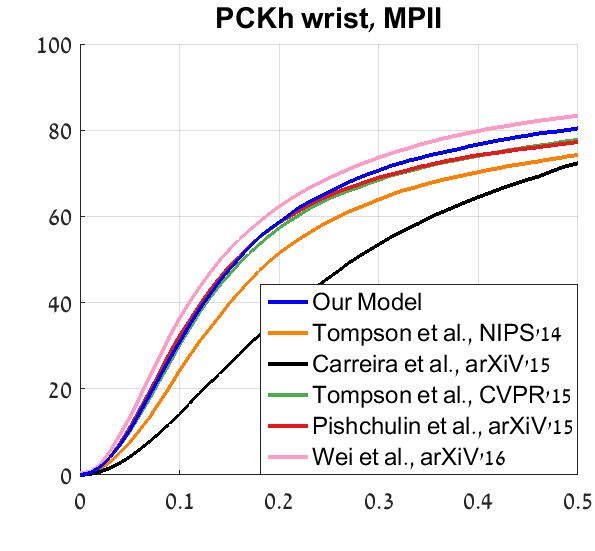}
	}%
	
	\caption{Pose estimation results on the MPII dataset for varying PCKh thresholds.}
	\label{fig:conditional-probability}
\end{figure}

\subsection{Leeds Sports}
The Leeds sports dataset \cite{Johnson10} (LSP) contains $2,000$ images of people in various sport activities, $1,000$ for training and $1,000$ for testing. The task is to return 14 keypoints, the same keypoints as in the MPII dataset except for the thorax and pelvis.\\

\begin{figure}[!ht]
	\centering
	\subfigure{%
		\includegraphics[width=0.125\textwidth]{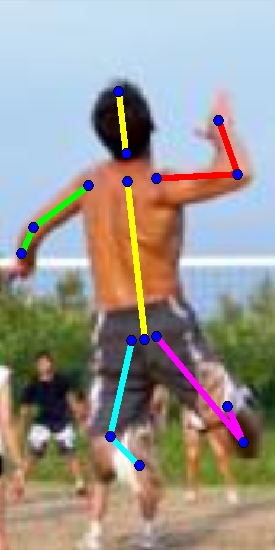}
	}%
	\subfigure{%
		\includegraphics[width=0.125\textwidth]{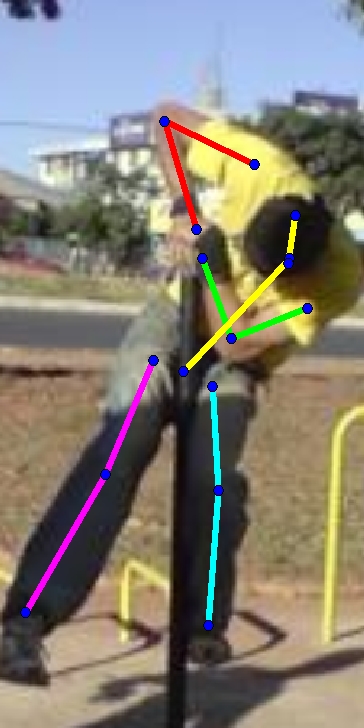}
	}%
	\subfigure{%
		\includegraphics[width=0.125\textwidth]{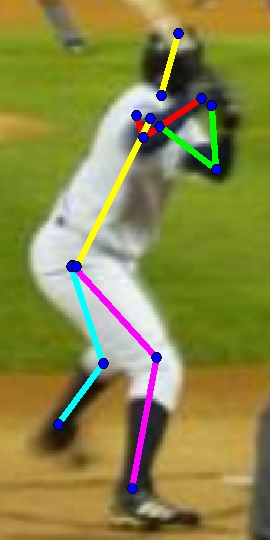}
	}%
	\subfigure{%
		\includegraphics[width=0.125\textwidth]{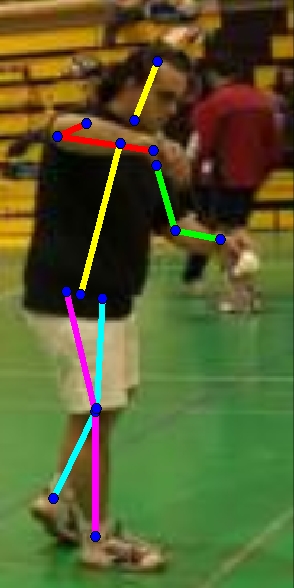}
	}%
	\subfigure{%
		\includegraphics[width=0.125\textwidth]{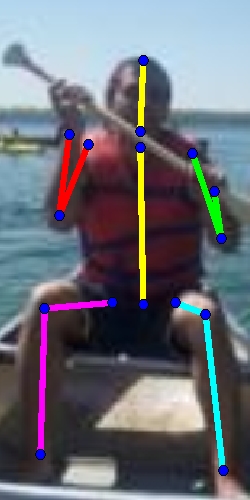}
	}%
	\subfigure{%
		\includegraphics[width=0.125\textwidth]{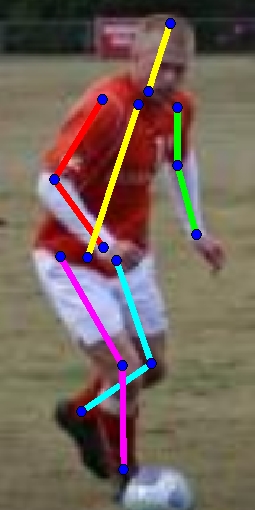}
	}%
	\subfigure{%
		\includegraphics[width=0.125\textwidth]{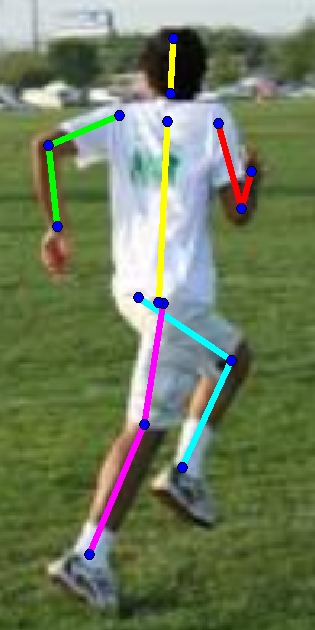}
	}
	
	\subfigure{%
		\includegraphics[width=0.18\textwidth]{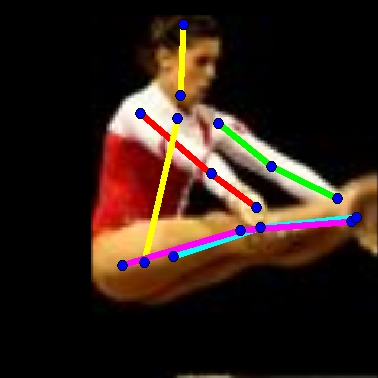}
	}%
	\subfigure{%
		\includegraphics[width=0.18\textwidth]{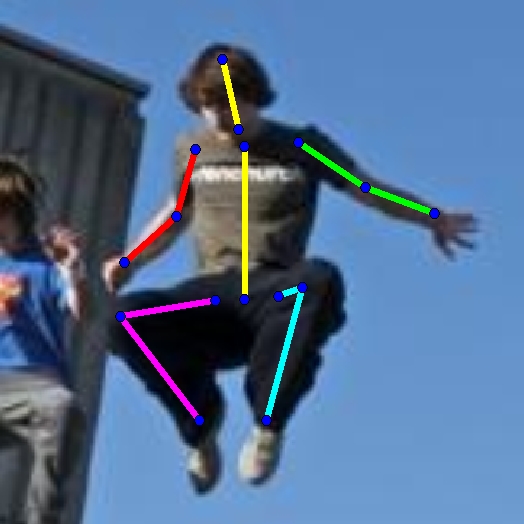}
	}%
	\subfigure{%
		\includegraphics[width=0.18\textwidth]{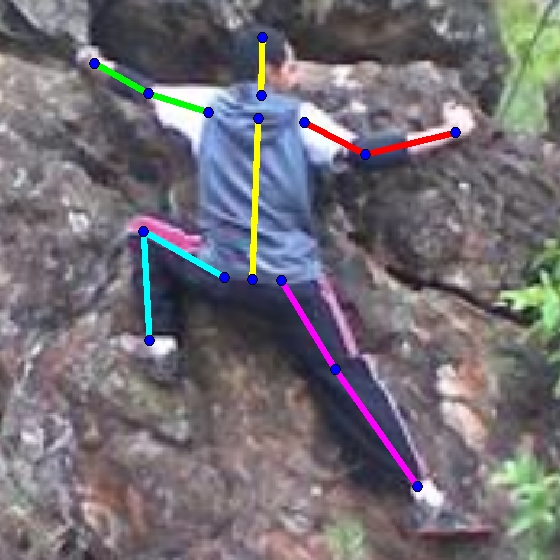}
	}%
	\subfigure{%
		\includegraphics[width=0.18\textwidth]{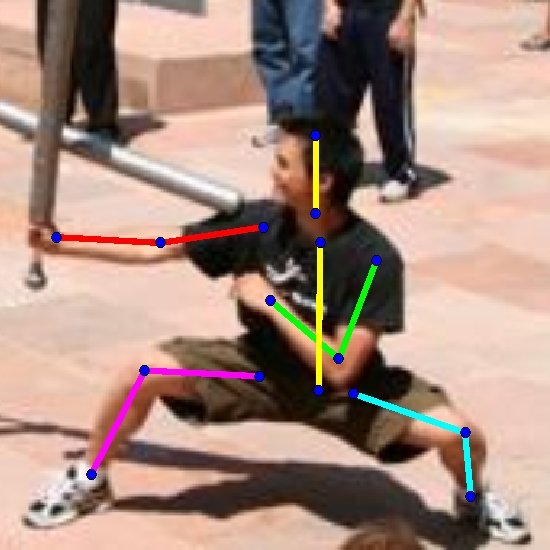}
	}%
	\subfigure{%
		\includegraphics[width=0.18\textwidth]{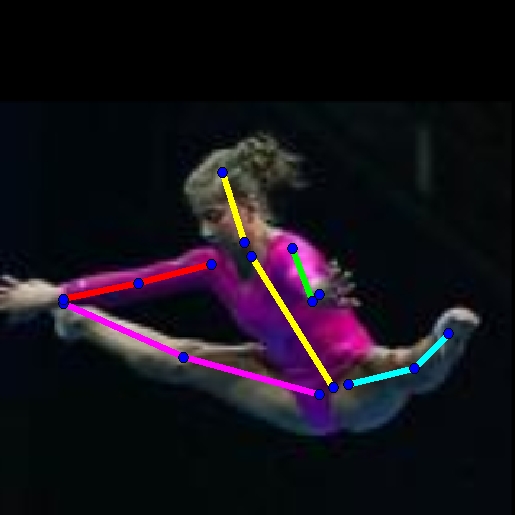}
	}%
	
	\caption{Additional results of our model's predicted joint positions on the LSP database test-set \cite{Johnson10}} 
	\label{fig:lsp-pred-images}
\end{figure}

The LSP dataset has two evaluation settings, person-centric (PC) and observer-centric (OC). We use the person-centric settings where right/left labels are according to body parts  (same as the MPII annotation) and not according to relative image location. The standard performance measure for the LSP dataset is strict percentage of correct parts (PCP) metric. The  PCP score measures limb detection: a limb is considered to be correctly detected if the distances between the detected limb endpoints and groundtruth limb endpoints are within half of the limb length. \\

\begin{table}[]

	\centering
	
	\label{my-label}
	\resizebox{\linewidth}{!}{%
	\begin{tabular}{|l|l|l|l|l|l|l|l|}
		
		\hline
		\textbf{Method} & \textbf{Torso} & \textbf{Upper} & \textbf{Lower} & \textbf{Upper} & \textbf{Forearm} & \textbf{Head} & \textbf{Mean} \\ 
		\textbf{} & \textbf{} & \textbf{Leg} & \textbf{Leg} & \textbf{Arm} & \textbf{} & \textbf{} & \textbf{} \\ \hline
		
		Tompson et al.\cite{Tompson14}& 90.3  & 70.4  & 61.1  & 63.0  & 51.2  & 83.7 & 66.6 \\  \hline
		
		Carreira et al.\cite{carreira15}& 95.3  & 81.8  & 73.3  & 66.7  & 51.0  & 84.4 & 72.5 \\  \hline

		Chen\&Yuille \cite{chen14}& 96.0  & 77.2  & 72.2  & 69.7  & 58.1  & 85.6 & 73.6 \\ \hline
		
		Fan et. al. \cite{Fan15}& 95.4  & 77.7  & 69.8  & 62.8  & 49.1  & 86.6 & 70.1 \\ \hline
		Pishchulin et al. \cite{pishchulin15}& 97.0  & 88.8  & 82.0  & \bf{82.4}  & \bf{71.8}  & \bf{95.8} & \bf{84.3} \\ \hline
		
		Our Model& \bf{97.3}  & \bf{88.9}  & \bf{84.5}  & 80.4  & 71.4  & 94.7 & 84.2 \\ \hline
	\end{tabular} 
}
	\caption{PCP results on the LSP dataset (PC). }

\end{table} 

We use the model trained on the MPII human pose dataset and fine-tune it on the LSP training set. Examples of our model's predictions are shown in fig. \ref{fig:lsp-pred-images}. At test time we run our algorithm twice, once with the input image and once with it flipped up-side-down, and pick the pose with the optimal score. This is done in order to handle up-side-down people which are more common in the LSP dataset than the MPII dataset, and are therefore under-represented at training time.\\

We compare the performance of our approach to that of other leading pose estimation methods in table 3. Our performance is comparable to that of Pishchulin et al. \cite{pishchulin15}, and  superior to other methods. We note that in \cite{pishchulin15} the authors use the LSP-Extended dataset, containing additional 10,000 annotated poses, not used in our model.
\section{Discussion}
In this work, we proposed a method for dealing with the challenging task of human pose estimation in still images. We presented a novel approach of using a deep convolutional neural net for keypoint voting rather than keypoint detection. The keypoint voting scheme has several useful properties compared to keypoint detection. First, all image regions of the evaluated person participate in the voting, utilizing the 'wisdom of the crowd' to produce reliable keypoint predictions. Second, any informative location can contribute to multiple keypoints, not just to a single one. This allows us to use consensus voting in order to compute expressive image-dependent joint keypoint probabilities. Empirical results on the diverse MPII human pose and Leeds sports pose datasets show competitive results, improving the state-of-the-art on a subset of evaluated keypoints. We showed that our model generalized well from the MPII dataset to the LSP dataset, using only 1000 samples for fine tuning. Models and code will be made publicly available. Additional contributions of the current scheme are the use of log-polar bins for location prediction rather than estimating L2 translations, and the use of convolutions for fast aggregation of votes from multiple locations.\\


The voting scheme is a natural tool for estimating the locations of unobserved body parts. In future work, we plan to harness this property for dealing with occlusions, resulting from closely interacting people, which are difficult to handle by existing schemes. In addition, we plan to combine our voting scheme with iterative methods, refining predictions by feeding the output of previous iterations as input to the network.\\

\bibliographystyle{plain}
\bibliography{deep_pose_bib}
\end{document}